\documentclass[conference]{IEEEtran}
\IEEEoverridecommandlockouts
\usepackage{cite}
\usepackage{amsmath,amssymb,amsfonts}
\usepackage{algorithm}
\usepackage{algorithmic}
\usepackage{graphicx}
\usepackage{textcomp}
\usepackage{xcolor}
\usepackage{placeins}
\def\BibTeX{{\rm B\kern-.05em{\sc i\kern-.025em b}\kern-.08em
    T\kern-.1667em\lower.7ex\hbox{E}\kern-.125emX}}

\usepackage[affil-it]{authblk}

\begin{document}

\title{Superpixel-Based QUBO for Scalable Quantum-Enhanced Medical Image Segmentation
}

\author[1]{Mohammad Chalhoub}
\author[1]{Mahdi Chehimi}
\author[2,3]{Laia Domingo}
\author[4]{Omar Alhussein}
\author[5,6]{Ahmed Farouk}
\author[5]{\\Saif Al-Kuwari}
\affil[1]{\small Department of Electrical and Computer Engineering, American University of Beirut, Beirut, Lebanon}
\affil[2]{\small Centre de Visió per Computador (CVC), Barcelona, Spain}
\affil[3]{\small Ingenii Inc., New York, USA}
\affil[4]{\small KU 6G Research Center, College of Computing
and Mathematical Sciences, Khalifa University, Abu Dhabi 127788, UAE}
\affil[5]{\small Qatar Center for Quantum Computing, College of Science and Engineering, Hamad Bin Khalifa University, Doha, Qatar}
\affil[6]{\small Department of Computer Science, Faculty of Computers and Artificial Intelligence, Hurghada University, Hurghada, Egypt}
\affil[ ]{\textit{\{mhc26, mc127\}@aub.edu.lb}, \textit{ldomingo@cvc.uab.cat}, \textit{omar.alhussein@ku.ac.ae}, \textit{\{ahsalem,smalkuwari\}@hbku.edu.qa}.}

\maketitle

\begin{abstract}
Quadratic unconstrained binary optimization (QUBO) has emerged as a powerful framework for medical computing problems. Binary decision variables naturally represent clinical choices, making QUBO formulations well-suited for quantum annealing hardware. However, a fundamental scalability challenge limits practical deployment: problem size grows
rapidly with input dimensionality, creating computational bottlenecks that restrict applications to simplified scenarios. This paper addresses this challenge through hierarchical problem reduction, as demonstrated in medical image segmentation, where pixel-level QUBO formulations create over 65,000 variables for a 256$\times$256 image, forcing existing approaches to downsample to 42$\times$42 resolution and discard 97\% of pixel information. A superpixel-based QUBO framework is proposed using simple linear iterative
clustering (SLIC) to group pixels into perceptually meaningful regions, then formulate segmentation as QUBO over a region adjacency graph (RAG) combining min-cut and smoothness objectives. Validation on INbreast mammography breast cancer images demonstrates a 4.2\% improvement in segmentation quality (mean IoU 0.76 vs 0.73) with 33.0$\times$ computational speedup (0.67s vs 21.97s) and a 97.3\% reduction in problem size (1764 to 48 variables), all achieved while processing full-resolution images rather than downsampled versions. The reduced problem size also fits well within current quantum annealer connectivity limits, removing the embedding overhead that has historically blocked direct deployment of pixel-level QUBO segmentation on quantum hardware.
\end{abstract}

\vspace{-0.3cm}
\section{Introduction}\vspace{-0.2cm}
Medical image segmentation is fundamental for computer-aided diagnosis and treatment 
planning, dividing images into meaningful regions that match anatomical structures 
or disease areas~\cite{wei2023quantum}. In mammography, accurate segmentation helps 
radiologists distinguish benign and malignant lesions. In brain imaging, precise 
tumor boundaries guide surgical planning. In cardiac imaging, chamber segmentation 
enables functional assessment. The accuracy of these segmentations directly affects 
clinical decisions, making segmentation quality a patient safety issue. However, 
manual segmentation by expert radiologists is time-consuming and subject to 
inter-observer variability. As medical imaging datasets continue to grow in scale 
and complexity, automated segmentation methods that deliver both accuracy and 
computational efficiency at scale become increasingly critical.

Image segmentation can be naturally expressed as graph optimization, where pixels 
or regions become nodes and edges encode similarity between neighbors. This 
graph-based view has proven to be successful in classical computer vision using methods 
such as normalized cuts and graph cuts~\cite{boykov2004experimental,felzenszwalb2004efficient}. 
The binary nature of segmentation, where each element belongs to the foreground or 
background, makes it naturally suited for quadratic unconstrained binary 
optimization (QUBO). QUBO formulations are particularly attractive in this context because they map directly onto quantum annealing hardware such as D-Wave systems, where binary spin variables and pairwise couplings are the native computational primitives. This positions QUBO-based segmentation as a natural target for a near-term quantum advantage in medical imaging~\cite{wei2023quantum}. Recent work has explored QUBO formulations for medical image 
segmentation, achieving quality comparable to supervised deep learning without 
labeled training data~\cite{domingo2024quantum}. Interactive QUBO segmentation 
guided by user-provided seeds has demonstrated practical deployment on quantum 
annealers~\cite{wang2024implementation}, while coupled reconstruction-segmentation 
approaches have addressed inverse CT imaging problems~\cite{jun2025quantum}. 
However, all of these methods operate at the pixel level, where each pixel becomes a 
binary variable in the QUBO formulation.

For pixel-level approaches, a typical $256 \times 256$ medical image creates more 
than 65,000 binary variables in the QUBO problem. To achieve tractability, existing work 
downsamples to $42 \times 42$ pixels~\cite{domingo2024quantum}, reducing to 1,764 
variables but discarding 97\% of pixel information. This aggressive downsampling 
loses important structural details and boundary information that are particularly problematic 
for clinical applications where fine anatomical features may be diagnostically 
relevant. The 1,764-variable QUBO is also still too large to embed efficiently on current quantum annealers without expensive minor-embedding overhead, which degrades solution quality~\cite{wang2024implementation}. The fundamental tension remains: pixel-level approaches either require 
severe downsampling with information loss or face prohibitively large problem sizes 
when processing full-resolution medical images.

Recent comprehensive reviews demonstrate the dominance of deep learning 
architectures for X-ray and mammography segmentation~\cite{xu2024advances,abueed2026automatic}, 
with U-Net variants becoming standard due to their ability to capture both local 
and global features. Although these supervised methods achieve high accuracy, they 
require extensive labeled datasets and substantial computational resources for 
training. Graph-based optimization methods, including QUBO formulations, offer an 
alternative unsupervised approach that does not require labeled training data, 
making them particularly attractive for medical applications where expert annotations are scarce and expensive, and unsupervised QUBO segmentation has already been shown to match U-Net-level accuracy at a fraction of the runtime~\cite{domingo2024quantum}.

Superpixel techniques offer a method for over-segmenting images, clustering pixels 
into visually consistent regions that respect natural boundaries. The SLIC 
algorithm has gained popularity for its computational effectiveness and boundary 
adherence~\cite{achanta2012slic}. Recent work has demonstrated superpixels' 
effectiveness specifically for breast imaging: mammographic tumor segmentation using 
shape-guided approaches~\cite{ali2024breast}, calcification detection~\cite{ren2020calcification}, 
and semantic classification of breast ultrasound~\cite{huang2020segmentation,daoud2019automatic}. 
Modern methods have integrated superpixels with deep learning for medical image 
segmentation pre-training~\cite{zeng2026supercl}, and classical segmentation 
pipelines widely adopt superpixels as a preprocessing step that reduces 
computational cost while preserving quality~\cite{stutz2018superpixels}. However, 
despite their widespread application in both classical and deep learning approaches, 
superpixels have not yet been incorporated into QUBO-based segmentation models.

By operating at the superpixel level rather than the pixel level, we achieve massive 
problem size reduction while maintaining access to full-resolution image information. 
A $256 \times 256$ medical image contains 65,536 pixels but can be represented by tens to a few hundred superpixels generated from full-resolution data, a reduction of more than two orders of magnitude in the number of QUBO variables. This differs fundamentally from downsampling to $42 \times 42$ 
pixels: both reduce optimization variables, but superpixels preserve structural 
detail by analyzing all pixels during grouping, whereas downsampling permanently 
discards information before optimization begins. This distinction is critical for 
medical imaging, where subtle intensity variations and fine boundaries carry 
diagnostic significance. While superpixels have proven effective in classical 
segmentation and QUBO has shown promise for medical imaging, these approaches have 
remained separate. All existing QUBO-based segmentation methods operate at the 
pixel level, inheriting severe scalability limitations. To the best of our knowledge, our work is the first to 
combine superpixel representation with a QUBO formulation for medical image segmentation. In particular, this paper makes three key contributions:

\begin{itemize}
\item We develop the first superpixel-based QUBO formulation for medical 
image segmentation, reducing the optimization variables from thousands to 
tens (97.3\% reduction, 1764 to 48 variables on average) while preserving the full-resolution image information 
and boundary accuracy.

\item We demonstrate simultaneous improvements in both segmentation quality 
and computational efficiency on INbreast mammography: 4.2\% higher IoU 
(0.76 vs 0.73) with 33.0$\times$ speedup compared to 
pixel-level baseline while bringing the QUBO problem within the direct-embedding capacity of present-day quantum annealers.

\item We establish general principles for hierarchical problem reduction, identifying 
meaningful atomic units, constructing appropriate similarity measures, and 
balancing problem size against solution quality, applicable to other QUBO-based 
medical optimization problems, including feature selection and treatment planning.
\end{itemize}

The remainder of this paper is organized as follows. Section~\ref{sec:PROPOSED APPROACH} 
presents our superpixel-based QUBO formulation and complete segmentation pipeline. 
Section~\ref{sec:EXPERIMENTAL RESULTS} reports the experimental results on INbreast mammography 
images. Section~\ref{sec:CONCLUSION} concludes and discusses future directions.

\section{Proposed Approach}\label{sec:PROPOSED APPROACH}
\subsection{Problem Formulation}

We represent an image as an undirected weighted graph $G = (V, E, W)$ where nodes $V$ correspond to image elements (pixels or superpixels), edges $E$ connect spatially adjacent elements, and edge weights $W$ encode similarity between neighboring elements. Each binary variable $x_i$ indicates whether the node $i$ belongs to the foreground ($x_i = 1$) or background ($x_i = 0$). The segmentation task becomes finding the binary assignment that optimally partitions the graph with respect to both boundary contrast and regional coherence.

The energy function for segmentation combines two complementary objectives. The min-cut term encourages placing the segmentation boundary along edges with low similarity (high contrast), while the smoothness term promotes spatial coherence by penalizing isolated or fragmented regions. We express this as ~\cite{domingo2024quantum}:
\begin{equation}\small
E(\mathbf{x}) = \sum_{(i,j) \in E} w_{ij} x_i (1 - x_j) + \alpha \sum_{(i,j) \in E} w_{ij} (1 - \delta(x_i, x_j)),
\end{equation}
where $w_{ij}$ represents the similarity weight between adjacent nodes $i$ and $j$, $\delta(x_i, x_j)$ is the Kronecker delta (equals 1 if $x_i = x_j$, 0 otherwise), and $\alpha$ is a hyperparameter controlling the relative importance of smoothness. The first term penalizes cutting edges with high similarity, while the second term penalizes pairs of neighboring nodes with different labels.

For binary variables $x_i \in \{0,1\}$, the Kronecker delta admits the polynomial representation $\delta(x_i, x_j) = (x_i + x_j - 1)^2 = 1 - x_i - x_j + 2x_i x_j$. Expanding and collecting terms yield 
the standard QUBO form $E(\mathbf{x}) = \mathbf{x}^T Q \mathbf{x} + \mathbf{c}^T \mathbf{x}$ 
with coefficients:
\begin{equation}
\begin{cases}\label{eq2}
c_i = (2\alpha + 1)\sum_{j} W_{ij} \\
Q_{ij} = -2(1 + \alpha)W_{ij} \\
Q_{ii} = -\alpha\sum_{j} W_{ij}
\end{cases},
\end{equation}

For each node $i$, the linear coefficient $c_i$ accumulates contributions from 
all adjacent edges. For each edge $(i,j)$, the quadratic coefficient $Q_{ij}$ 
encodes the interaction between the nodes $i$ and $j$. The diagonal terms $Q_{ii}$ 
arise from the expansion of the Kronecker delta and contribute to the smoothness 
penalty. We retain $c_i$ and $Q_{ii}$ as separate quantities to match the standard input format expected by both classical simulated annealers and quantum annealing samplers, although $x_i^2 = x_i$ for binary variables. These coefficients are accumulated over all edges in the graph and can 
be directly inputted to quantum or classical annealers.

\subsection{Superpixel Generation and Feature Extraction}

The first stage of our approach generates superpixels from the full-resolution 
medical image using the SLIC (Simple Linear Iterative Clustering) algorithm~\cite{achanta2012slic}. 
SLIC performs $k$-means clustering in a five-dimensional space combining spatial 
coordinates and color channels. For grayscale medical images, we convert them to RGB 
format for compatibility with SLIC, then cluster pixels based on their spatial 
location $(x, y)$ coordinates and intensity values.

SLIC is controlled by two main parameters that determine the resulting superpixel 
characteristics. The number of segments parameter $n_{\text{segments}}$ specifies a target number of superpixels to generate, though the final count may be slightly smaller after small or disconnected regions are merged during post-processing. The algorithm initializes cluster 
centers on a regular grid with spacing $S = \sqrt{N/n_{\text{segments}}}$, where 
$N$ is the total number of pixels. For a $256 \times 256$ image with
$n_{\text{segments}} = 60$ as used in our experiments, this yields an initial cluster spacing of approximately
33 pixels. The compactness parameter controls the relative importance of spatial 
proximity versus intensity similarity in the clustering objective. Higher compactness 
values enforce more regular, square-shaped superpixels, while lower values allow 
superpixels to adapt more freely to intensity boundaries. We use 
$\text{compactness} = 10$ based on empirical validation showing good boundary 
adherence while maintaining computational efficiency~\cite{achanta2012slic}

The SLIC algorithm iterates between two steps until convergence. In the assignment 
step, each pixel is assigned to the closest cluster center based on a combined 
distance metric that weighs spatial and intensity differences. In the update step, 
cluster centers are recomputed as the mean position and intensity of all pixels 
assigned to each cluster. This process typically converges in 10--15 iterations, 
producing a label map where each pixel is assigned an integer superpixel ID.

After superpixel generation, we extract the features for each superpixel region. For 
each unique superpixel ID $k$, we compute the mean intensity $\mu_k$ by averaging 
all pixel intensities within that superpixel:
\begin{equation}
\mu_k = \frac{1}{|P_k|} \sum_{p \in P_k} I(p),
\end{equation}
where $P_k$ is the set of pixels belonging to superpixel $k$, $|P_k|$ is the 
number of pixels in that set, and $I(p)$ is the intensity value at pixel $p$. 
This mean intensity serves as the primary feature representing each superpixel 
in subsequent optimization stages.

\begin{algorithm}[t]
\caption{Superpixel-Based QUBO Image Segmentation}
\label{alg:pipeline_algorithm}
\small
\begin{algorithmic}[1]
\STATE \textbf{Input:} Medical image $I$ of size $H \times W$
\STATE \textbf{Input:} Number of superpixels $K$, compactness $c$, smoothness $\alpha$, runs $R$
\STATE \textbf{Output:} Binary segmentation mask $M$ of size $H \times W$
\STATE
\STATE \textbf{// Preprocessing}
\STATE $I \gets \text{ContrastEnhancement}(I)$
\STATE
\STATE \textbf{// Superpixel Generation}
\STATE $L \gets \text{SLIC}(I, n_{\text{segments}}=K, \text{compactness}=c)$
\STATE
\STATE \textbf{// Feature Extraction}
\FOR{each unique superpixel ID $k$ in $L$}
    \STATE $P_k \gets \{p : L(p) = k\}$
    \STATE $\mu_k \gets \text{mean}(\{I(p) : p \in P_k\})$
\ENDFOR
\STATE
\STATE \textbf{// Region Adjacency Graph Construction}
\STATE $E \gets \text{BuildRAG}(L)$ \COMMENT{Scan right/down for adjacency}
\STATE
\STATE \textbf{// Weight Computation}
\STATE $\sigma \gets \text{StandardDeviation}(I)$
\FOR{each edge $(k,l) \in E$}
    \STATE $w_{kl} \gets \exp\left(-\frac{(\mu_k - \mu_l)^2}{2\sigma^2}\right)$
    \STATE $w_{kl} \gets -1 \times (1 - w_{kl})$
\ENDFOR
\STATE $W \gets \text{MinMaxNormalize}(\{w_{kl}\}, \text{range}=[-1,1])$
\STATE
\STATE \textbf{// Build Weighted Graph}
\STATE $G \gets (V, E, W)$ where $V = \{1, 2, \ldots, K\}$
\STATE
\STATE \textbf{// QUBO Construction}
\STATE $c, Q \gets \text{ConstructQUBO}(G, \alpha)$ \COMMENT{Using Equation (2)}
\STATE
\STATE \textbf{// Optimization}
\STATE $\mathbf{x}^* \gets \text{SimulatedAnnealing}(c, Q, \text{runs}=R)$
\STATE
\STATE \textbf{// Pixel-Level Reconstruction}
\STATE $M \gets \text{zeros}(H, W)$
\FOR{each superpixel ID $k$}
    \STATE $M[L == k] \gets \mathbf{x}^*[k]$
\ENDFOR
\STATE
\STATE \textbf{return} $M$
\end{algorithmic}
\end{algorithm}

\begin{figure*}[t]
\centering
\includegraphics[width=\textwidth]{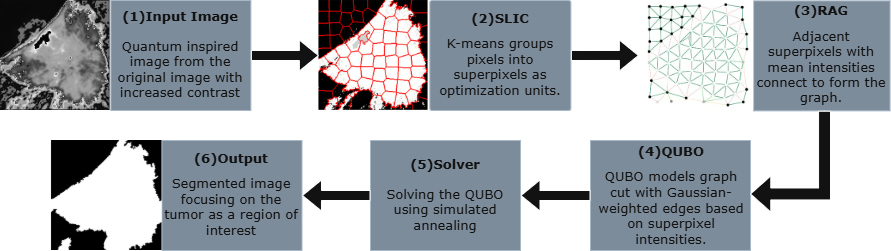}
\caption{Superpixel-based QUBO segmentation pipeline. (1) Input mammography image undergoes quantum-inspired contrast-enhancement preprocessing. (2) SLIC algorithm generates ~50 superpixels from 65,536 pixels. (3) Region Adjacency Graph connects adjacent superpixels with similarity-weighted edges. (4) QUBO formulation models segmentation as a graph cut with Gaussian weights. (5) Simulated annealing optimizes the binary labeling. (6) Full-resolution segmentation mask isolates the tumor region. The pipeline reduces problem complexity by 97.3\% while maintaining image fidelity.}\vspace{-0.5cm}
\label{fig:pipeline}
\end{figure*}

\subsection{Region Adjacency Graph and Weight Computation}

The Region Adjacency Graph (RAG) represents spatial relationships between 
superpixels as an undirected graph where nodes correspond to superpixels and 
edges connect spatially adjacent regions~\cite{achanta2012slic}. We construct 
the RAG by scanning the superpixel label map and identifying boundary pixels 
where neighboring superpixels meet.

Our RAG construction algorithm processes the label map in raster order, examining 
each pixel's right and down neighbors to detect adjacency. For each pixel at 
position $(i, j)$ with superpixel label $L(i, j)$, we check if the pixel to the 
right at $(i, j+1)$ has a different label. If so, we record an adjacency between 
the superpixels $L(i, j)$ and $L(i, j+1)$. Similarly, we check the pixel below at 
$(i+1, j)$. This scanning approach ensures that we identify all adjacencies while 
avoiding duplicates, as each boundary is encountered exactly once when scanning 
right and down. In our experimental setting with roughly 50 superpixels per image, the RAG contains on the order of 100 edges, reflecting that superpixels form a roughly planar graph in which each region typically touches 3--5 neighbors.

Edge weights in the RAG encode the similarity between adjacent superpixels, 
determining which boundaries the optimization should preserve or cut. We compute 
weights using Gaussian similarity based on mean superpixel intensities, following 
the approach established in~\cite{domingo2024quantum} for 
pixel-level segmentation but adapted to operate on superpixel features.

For each edge $(k,l) \in E$ connecting adjacent superpixels $k$ and $l$, we 
compute the raw similarity weight using the Gaussian kernel:
\begin{equation}
w_{kl}^{(\text{raw})} = \exp\left(-\frac{(\mu_k - \mu_l)^2}{2\sigma^2}\right),
\end{equation}
where $\mu_k$ and $\mu_l$ are the mean intensities of superpixels $k$ and $l$ 
respectively, and $\sigma$ is a scale parameter controlling sensitivity to intensity 
differences. We set $\sigma$ equal to the standard deviation of pixel intensities 
across the entire image, computed as:
\begin{equation}
\sigma = \sqrt{\frac{1}{N} \sum_{i=1}^{N} (I_i - \bar{\mu})^2},
\end{equation}
where $N$ is the total number of pixels, $I_i$ is the intensity at pixel $i$, 
and $\bar{\mu}$ is the global mean intensity. This automatic scaling adapts the 
similarity measure to each image's contrast characteristics.

The raw Gaussian similarity produces values in the range $[0, 1]$, with values 
near 1 indicating very similar superpixels and values near 0 indicating dissimilar 
regions. To match the energy minimization framework where low energy corresponds 
to good segmentation, we transform the similarity to a cost: $w_{kl}^{(\text{cost})} = -1 \times (1 - w_{kl}^{(\text{raw})})$.

This transformation maps high similarity (near 1) to low cost (near 0), and low 
similarity (near 0) to high cost (near $-1$). Finally, the edge weights are min-max normalized to the range $[-1, 1]$ across the RAG to ensure the numerical stability of the QUBO solver and to prevent any single edge from dominating the optimization.




\subsection{QUBO Formulation and Optimization Pipeline}

Algorithm~\ref{alg:pipeline_algorithm} presents the complete superpixel-based QUBO 
segmentation pipeline, integrating all stages from preprocessing through final 
mask reconstruction. Fig.~\ref{fig:pipeline} illustrates the same pipeline visually. The QUBO construction step computes linear and quadratic coefficients using 
\eqref{eq2}. For each edge $(k,l)$ with weight $w$, we accumulate: 
$c[k] \mathrel{+}= (1 + 2\alpha)w$, $c[l] \mathrel{+}= (1 + 2\alpha)w$, 
$Q[k,l] \mathrel{+}= -2(1 + \alpha)w$, $Q[k,k] \mathrel{+}= -\alpha w$, 
and $Q[l,l] \mathrel{+}= -\alpha w$.

We used simulated annealing with a linear cooling schedule from $T=0.1$ to 
$T=4.2$ over 2000 iterations. The smoothness parameter is set to $\alpha=10$ ~\cite{domingo2024quantum}. The solver returns a binary label assignment 
for each superpixel. We map these labels back to the pixel level by setting all 
pixels within the superpixel $k$ to the value of $\mathbf{x}^*[k]$. This produces 
the final segmentation mask at the original image resolution, completing the 
pipeline from raw medical image to binary segmentation without lossy downsampling.

\begin{table*}[t]
\centering
\caption{Average Performance Across 40 Medical Images}
\label{tab:average}
\small
\begin{tabular}{lccccc}
\hline
\textbf{Method} & \textbf{IoU} & \textbf{Dice} & \textbf{Nodes} & \textbf{Edges} & \textbf{Time (s)} \\
 & (mean $\pm$ std) & (mean $\pm$ std) & (mean) & (mean) & (mean $\pm$ std) \\
\hline
Pixel Gaussian & 0.73 $\pm$ 0.14 & 0.83 $\pm$ 0.1 & 1764.0 & 3444.0 & 21.97 $\pm$ 10.64 \\
SP Gaussian & \textbf{0.76 $\pm$ 0.14} & \textbf{0.86 $\pm$ 0.1} & \textbf{47.8} & \textbf{118.0} & \textbf{0.67 $\pm$ 0.42} \\
\hline
Improvement & +4.2\% & +2.4\% & -97.3\% & -96.6\% & -97.0\% \\
\hline
\end{tabular}
\end{table*}

\section{ExperimentalResults}\label{sec:EXPERIMENTAL RESULTS}
\subsection{Experimental Setup}

We evaluated our superpixel-based QUBO segmentation approach on 40 mammography 
images from the INbreast dataset~\cite{moreira2012inbreast}. These images correspond to all cases in the dataset that include both expert lesion annotations and the preprocessing format used by the pixel-level baseline of~\cite{domingo2024quantum}, ensuring a fair comparison on identical inputs. The following parameters are used throughout our experiments: 1) Superpixel generation: SLIC algorithm~\cite{achanta2012slic} with 
compactness $c = 10$, 2) Number of segments: We tuned $K$ on a held-out subset by sweeping $K \in \{15, 25, 60, 100, 200\}$ and selected the value achieving the best mean IoU across the dataset, which was $K = 60$. This single global value is then applied to all 40 images in the reported comparison, 3) QUBO formulation: Smoothness parameter $\alpha = 10$, 4) Optimization: Simulated annealing with 2000 iterations, 5) Baseline: Pixel-level QUBO on $42 \times 42$ downsampled images 
(1764 variables)~\cite{domingo2024quantum}, and 6) Our approach: Superpixel-based QUBO on full $256 \times 256$ images. Additionally, segmentation quality is measured using two standard overlap metrics computed 
after automatic polarity correction:
\begin{equation}
\text{IoU} = \frac{|P \cap G|}{|P \cup G|}, \quad
\text{Dice} = \frac{2|P \cap G|}{|P| + |G|},
\end{equation}
where $P$ is the predicted segmentation and $G$ is the ground truth mask. Both 
metrics range from 0 (no overlap) to 1 (perfect agreement)~\cite{taha2015metrics}. 
Computational efficiency is measured by the total pipeline runtime, including 
preprocessing, graph construction, QUBO formulation, and optimization.

\subsection{Average Performance Across 40 Images}

Table~\ref{tab:average} summarizes results across 40 mammography images, with $K=60$ used uniformly for the superpixel method. Using a single globally selected $K$ rather than per-image tuning ensures that the reported comparison reflects an out-of-the-box deployment scenario rather than oracle-tuned performance

The superpixel method achieves a mean IoU of 0.76 $\pm$ 0.14 compared to 
0.73 $\pm$ 0.14 for the pixel-based baseline, a 4.2\% relative improvement. Mean Dice 
coefficient improves from 0.83 $\pm$ 0.1 to 0.86 $\pm$ 0.1 (+2.4\%). The improvements are consistent across the 40-image cohort: the superpixel method outperformed the pixel-level baseline on the majority of images, with the largest gains observed for lesions whose boundaries align poorly with the $42 \times 42$ downsampling grid.

The average computational time decreases from 21.97 $\pm$ 10.64 seconds to 
0.67 $\pm$ 0.42 seconds, representing a 33.0$\times$ mean speedup. The remaining variability in superpixel runtime is driven by image-dependent variation in the actual number of superpixels produced and the resulting RAG density, rather than by the choice of $K$, which is kept fixed at 60. Even the slowest superpixel runs remain at least an order of magnitude faster than the pixel-level baseline while delivering superior segmentation quality.

\subsection{Problem Size Reduction Analysis}

Table~\ref{tab:reduction} quantifies the dramatic reduction in QUBO complexity 
achieved through superpixels. On average, the number of binary 
variables decreases by 97.3\% (from 1764 to 48), and the edges decrease by 96.6\% 
(from 3444 to 118). The QUBO matrix size, which scales as the square of the 
number of variables, reduces by 99.9\%, from 3.1 million potential entries to 
just 2.3 thousand. This reduction is what enables practical optimization on 
classical and quantum hardware.

\begin{table}[t]
\centering
\caption{QUBO Problem Size Reduction Through Superpixels}
\label{tab:reduction}
\setlength{\tabcolsep}{4pt} 
\begin{tabular}{lccc}
\hline
\textbf{Metric} & \textbf{Pixel-Level} & \textbf{Superpixel-Level} & \textbf{Reduction} \\
\hline
Binary Variables & 1764 & $47.8 \approx 48$ & 97.3\% \\
Graph Edges & 3444 & $117.95 \approx 118$ & 96.6\% \\
QUBO Matrix Size & $1764^2 = 3.1$M & $47.8^2 = 2.28$K & 99.9\% \\
Optimization Time (s) & 21.97 $\pm$ 10.64 & 0.67 $\pm$ 0.42 & 97.0\% \\
\hline
Speedup Factor & \multicolumn{3}{c}{33.0$\times$ average} \\
\hline
\end{tabular}\vspace{-0.4cm}
\end{table}

For quantum hardware deployment, this reduction is particularly critical. Current 
quantum annealers like D-Wave Advantage have approximately 5000 qubits but face 
strict connectivity constraints that limit direct problem mapping. A pixel-level 
QUBO with 1764 variables would require expensive minor embedding that degrades 
solution quality. In contrast, a superpixel-level QUBO with roughly 50 variables can 
be embedded with minimal overhead and, in many cases, mapped almost directly to the hardware graph. This makes superpixel-based 
formulation of a practical pathway for deploying quantum-enhanced medical image 
segmentation in clinical settings. Critically, this problem size reduction is achieved while processing full-resolution 
images rather than downsampled versions, preserving the clinical utility of the 
segmentation for diagnostic applications.

\begin{figure*}[t]
\centering
\includegraphics[width=\textwidth]{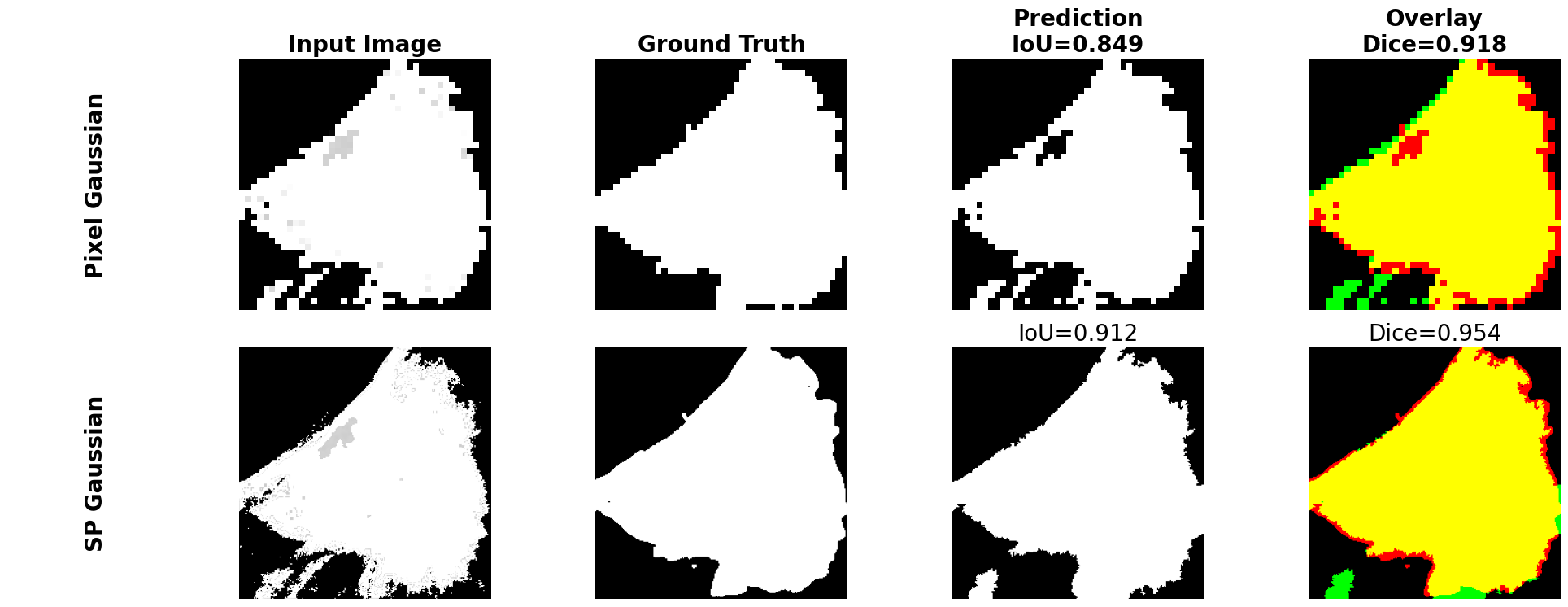}
\caption{Visual comparison of pixel-level and superpixel-level QUBO segmentation 
on mammography image (idx=7). Top row: Pixel-level baseline on $42 \times 42$ 
downsampled image shows blocky artifacts. Bottom row: Superpixel approach on 
full $256 \times 256$ resolution preserves fine anatomical details. Overlay 
shows correct predictions (yellow), false positives (red), and false negatives 
(green).}\vspace{-0.5cm}
\label{fig:visual}
\end{figure*}

\subsection{Discussion}
Our results indicate that aggressive parameter reduction need not come at the cost of accuracy when the reduction is performed along task-irrelevant dimensions. Although our formulation has 97.3\% fewer free variables than the pixel-level baseline, the discarded degrees of freedom correspond to sub-superpixel decisions inside visually uniform regions, where segmentation boundaries are unlikely to lie. Superpixels therefore act as a structural prior that removes variables the QUBO does not need, rather than information the task does, fundamentally different from downsampling which discards content indiscriminately. The resulting 4.2\% IoU improvement alongside a 33$\times$ speedup is consistent with this reduction rather than in spite of it: a task-aligned inductive bias substitutes for raw representational capacity. More broadly, the atomic units of optimization need not be the atomic units of the input data; aligning problem granularity with perceptually meaningful regions matches the optimization to the underlying clinical task.

The visual comparison in Figure~\ref{fig:visual} reveals why superpixels 
outperform pixel-level approaches. Pixel-level errors are uniformly distributed 
along boundaries, reflecting the fundamental limitation that downsampling cannot 
distinguish genuine boundary uncertainty from resolution artifacts. Superpixel 
errors concentrate in ambiguous regions where intensity gradients are weak, precisely 
the regions where segmentation is genuinely difficult. This suggests that 
superpixel-based QUBO is approaching the intrinsic difficulty of the segmentation 
task rather than being limited by representational constraints.

QUBO-based segmentation has been shown to achieve accuracy comparable to supervised U-Net architectures on this task while replacing minutes-long supervised training on annotated datasets with sub-second inference on quantum or quantum-inspired solvers~\cite{domingo2024quantum}. Our superpixel-based formulation strengthens this trade-off in two ways. First, it preserves the unsupervised nature of the QUBO approach, requiring no labeled data, while raising mean IoU by 4.2\% over the pixel-level QUBO baseline. Second, reducing the problem to roughly 48 binary variables brings the end-to-end runtime down to 0.67 seconds on a classical simulated annealer, eliminating the embedding overhead that has historically been the bottleneck for quantum annealing of pixel-level QUBOs. The result is a segmentation pipeline that is competitive in quality with supervised deep learning, faster than U-Net training, and well-matched in structure to quantum annealing hardware.

However, our evaluation has several limitations. We compared Gaussian similarity against mutual information and bilateral filtering on mammography, and Gaussian similarity on mean superpixel intensity consistently performed best, likely because mammograms are single-channel and the well-summarized mean intensity provides little additional signal for higher-order joint-distribution measures, while the QUBO smoothness term already absorbs the spatial-coherence role of bilateral weighting. These conclusions may not transfer to multi-channel modalities such as multi-parametric MRI. The optimal number of superpixels $K$ varies across images, requiring either 
manual tuning or automated selection procedures. Extending to multi-class 
segmentation would require modifications to the binary QUBO formulation. The cohort size of 40 images, while consistent with prior QUBO segmentation studies~\cite{domingo2024quantum}, is modest by deep-learning standards and a larger-scale evaluation across modalities is needed to characterize the approach more fully. Finally, all reported results use classical simulated annealing; while the reduced problem size makes direct quantum-annealer deployment feasible, an empirical comparison on real quantum hardware is left for future work. Despite these limitations, the hierarchical reduction principle demonstrated 
here, identifying meaningful atomic units and formulating optimization at the 
appropriate granularity, applies broadly to other medical optimization problems 
where computational scalability currently limits practical deployment.

\section{Conclusion}\label{sec:CONCLUSION}
Pixel-level QUBO formulations for medical image segmentation force aggressive 
downsampling that discards most of the pixel information. We addressed this 
through superpixel-based hierarchical reduction, formulating QUBO over a 
Region Adjacency Graph of perceptually meaningful regions rather than individual 
pixels. On INbreast mammography, the proposed method improves mean IoU from 
0.73 to 0.76 (+4.2\%) while reducing problem size from 1,764 to 48 variables 
(97.3\%) and runtime from 21.97 to 0.67 seconds (33$\times$ speedup), all while 
processing full-resolution images. The reduced problem size also fits within the 
direct-embedding capacity of present-day quantum annealers. Future work will 
validate the formulation on D-Wave hardware, extend it to CT and MRI, and apply 
the same hierarchical-reduction principle to other QUBO-based clinical problems 
such as genomic feature selection and radiation treatment planning.

\section*{Acknowledgment}
This work was supported by the Vertical Integrated Project at the American University of Beirut.

\bibliographystyle{IEEEtran} 
\bibliography{references}

@article{wei2023quantum,
  title={Quantum machine learning in medical image analysis: A survey},
  author={Wei, Lin and Liu, Haowen and Xu, Jing and Shi, Lei and Shan, Zheng and Zhao, Bo and Gao, Yufei},
  journal={Neurocomputing},
  volume={525},
  pages={42--53},
  year={2023},
  publisher={Elsevier}
}

@article{domingo2024quantum,
  title={Quantum-enhanced unsupervised image segmentation for medical images analysis},
  author={Domingo, Laia and Chehimi, Mahdi},
  journal={arXiv preprint arXiv:2411.15086},
  year={2024}
}

@article{wang2024implementation,
  title={Implementation and analysis of quantum-classical hybrid interactive image segmentation algorithm based on quantum annealer: K. Wang et al.},
  author={Wang, Kehan and Wang, Shuang and Chen, Qinghui and Qiao, Xingyu and Ma, Hongyang and Qiu, Tianhui},
  journal={Quantum Information Processing},
  volume={23},
  number={8},
  pages={301},
  year={2024},
  publisher={Springer}
}

@article{jun2025quantum,
  title={Quantum optimization algorithms for CT image segmentation from X-ray data},
  author={Jun, Kyungtaek and Lee, Hyunju},
  journal={Scientific Reports},
  volume={15},
  number={1},
  pages={20649},
  year={2025},
  publisher={Nature Publishing Group UK London}
}

@article{achanta2012slic,
  title={SLIC superpixels compared to state-of-the-art superpixel methods},
  author={Achanta, Radhakrishna and Shaji, Appu and Smith, Kevin and Lucchi, Aurelien and Fua, Pascal and S{\"u}sstrunk, Sabine},
  journal={IEEE transactions on pattern analysis and machine intelligence},
  volume={34},
  number={11},
  pages={2274--2282},
  year={2012},
  publisher={IEEE}
}

@article{stutz2018superpixels,
  title={Superpixels: An evaluation of the state-of-the-art},
  author={Stutz, David and Hermans, Alexander and Leibe, Bastian},
  journal={Computer Vision and Image Understanding},
  volume={166},
  pages={1--27},
  year={2018},
  publisher={Elsevier}
}

@article{boykov2004experimental,
  title={An experimental comparison of min-cut/max-flow algorithms for energy minimization in vision},
  author={Boykov, Yuri and Kolmogorov, Vladimir},
  journal={IEEE transactions on pattern analysis and machine intelligence},
  volume={26},
  number={9},
  pages={1124--1137},
  year={2004},
  publisher={IEEE}
}

@article{felzenszwalb2004efficient,
  title={Efficient graph-based image segmentation},
  author={Felzenszwalb, Pedro F and Huttenlocher, Daniel P},
  journal={International journal of computer vision},
  volume={59},
  number={2},
  pages={167--181},
  year={2004},
  publisher={Springer}
}

@article{moreira2012inbreast,
  title={Inbreast: toward a full-field digital mammographic database},
  author={Moreira, In{\^e}s C and Amaral, Igor and Domingues, In{\^e}s and Cardoso, Ant{\'o}nio and Cardoso, Maria Joao and Cardoso, Jaime S},
  journal={Academic radiology},
  volume={19},
  number={2},
  pages={236--248},
  year={2012},
  publisher={Elsevier}
}

@article{taha2015metrics,
  title={Metrics for evaluating 3D medical image segmentation: analysis, selection, and tool},
  author={Taha, Abdel Aziz and Hanbury, Allan},
  journal={BMC medical imaging},
  volume={15},
  number={1},
  pages={29},
  year={2015},
  publisher={Springer}
}

@article{xu2024advances,
  title={Advances in medical image segmentation: A comprehensive review of traditional, deep learning and hybrid approaches},
  author={Xu, Yan and Quan, Rixiang and Xu, Weiting and Huang, Yi and Chen, Xiaolong and Liu, Fengyuan},
  journal={Bioengineering},
  volume={11},
  number={10},
  pages={1034},
  year={2024},
  publisher={MDPI}
}

@article{abueed2026automatic,
  title={Automatic semantic segmentation in chest X-ray images using deep learning approaches: a literature review},
  author={Abueed, Omar and Thakkar, Priyank and AlAlaween, Wafa’H and Wang, Yong and Khasawneh, Mohammad T},
  journal={Neural Computing and Applications},
  volume={38},
  number={4},
  pages={70},
  year={2026},
  publisher={Springer}
}

@article{ali2024breast,
  title={Breast tumor segmentation using neural cellular automata and shape guided segmentation in mammography images},
  author={Ali, Mudassar and Wu, Tong and Hu, Haoji and Mahmood, Tariq},
  journal={Plos one},
  volume={19},
  number={10},
  pages={e0309421},
  year={2024},
  publisher={Public Library of Science San Francisco, CA USA}
}

@article{ren2020calcification,
  title={Calcification segmentation based on a different scales superpixels saliency detection algorithm},
  author={Ren, Li and Liu, Yangyang and Tong, Ying and Cao, Xuehong and Wu, Yiyun},
  journal={Ultrasound in Medicine \& Biology},
  volume={46},
  number={12},
  pages={3404--3412},
  year={2020},
  publisher={Elsevier}
}

@article{huang2020segmentation,
  title={Segmentation of breast ultrasound image with semantic classification of superpixels},
  author={Huang, Qinghua and Huang, Yonghao and Luo, Yaozhong and Yuan, Feiniu and Li, Xuelong},
  journal={Medical Image Analysis},
  volume={61},
  pages={101657},
  year={2020},
  publisher={Elsevier}
}

@article{daoud2019automatic,
  title={Automatic superpixel-based segmentation method for breast ultrasound images},
  author={Daoud, Mohammad I and Atallah, Ayman A and Awwad, Falah and Al-Najjar, Mahasen and Alazrai, Rami},
  journal={Expert Systems with Applications},
  volume={121},
  pages={78--96},
  year={2019},
  publisher={Elsevier}
}

@article{zeng2026supercl,
  title={Supercl: Superpixel guided contrastive learning for medical image segmentation pre-training},
  author={Zeng, Shuang and Zhu, Lei and Zhang, Xinliang and He, Hangzhou and Lu, Yanye},
  journal={IEEE Transactions on Image Processing},
  year={2026},
  publisher={IEEE}
}

\end{document}